\documentclass[10pt, a4paper]{svproc}
\usepackage{url}

\usepackage{mathtools,amssymb,commath}
\usepackage{amsmath}
\usepackage{graphicx,import}
\usepackage{verbatim}
\usepackage{color}
\usepackage{hyperref}
\hypersetup{colorlinks=true}  
\usepackage{subcaption}
\usepackage{tikz}
\usepackage[linesnumbered,ruled]{algorithm2e}
\usepackage{algorithmic} 
\usepackage{booktabs}

\DeclareMathOperator*{\argmax}{arg\,max}
\DeclareMathOperator*{\argmin}{arg\,min}

\usepackage{mathtools,amsthm}

\begin{document}
\mainmatter              
\title{Nonlinear Stochastic Trajectory Optimization for Centroidal Momentum Motion Generation of Legged Robots}
\titlerunning{Nonlinear Stochastic Trajectory Optimization}  
%
\author{Ahmad Gazar\inst{1} \and Majid Khadiv\inst{1}
\and Sébastien Kleff\inst{2,4} \and Andrea Del Prete\inst{3} \and
Ludovic Righetti\inst{1,4}}
\authorrunning{Ahmad Gazar et al.} 
%
%
\institute{Max Planck Institute for Intelligent Systems, Tuebingen, Germany,\\
\email{ahmad.gazar@tuebingen.mpg.de, majid.khadiv@tuebingen.mpg.de},\\
\and
LAAS-CNRS, Université de Toulouse, CNRS, Toulouse,\\
\and
Industrial Engineering Department, University of Trento, Italy,\\
\email{andrea.delprete@unitn.it},
\and
Tandon School of Engineering, New York University, New York, USA,\\
\email{sk8001@nyu.edu, ludovic.righetti@nyu.edu}}

\maketitle              
\begin{abstract}
Generation of robust trajectories for legged robots remains a challenging task due to the underlying nonlinear, hybrid and intrinsically unstable dynamics which needs to be stabilized through limited contact forces. Furthermore, disturbances arising from unmodelled contact interactions with the environment and model mismatches can hinder the quality of the planned trajectories leading to unsafe motions. In this work, we propose to use stochastic trajectory optimization for generating robust centroidal momentum trajectories to account for additive uncertainties on the model dynamics and parametric uncertainties on contact locations. Through an alternation between the robust centroidal and whole-body trajectory optimizations, we generate robust momentum trajectories while being consistent with the whole-body dynamics. We perform an extensive set of simulations subject to different uncertainties on a quadruped robot showing that our stochastic trajectory optimization problem reduces the amount of foot slippage for different gaits while achieving better performance over deterministic planning. 
\end{abstract}
\section{Introduction}
\label{sec:intro}
Trajectory optimization has become a dominant paradigm for planning and control of whole-body trajectories for legged robots \cite{mordatch2012,posa2014,winkler2018,carpentier2021}. Despite recent advancements in real-time trajectory re-planning \cite{sleiman2021unified,meduri2022biconmp}, dealing with uncertainties and assessing risk remains an open challenge in controlling legged robots. Uncertainties arising from inaccurate modelling, estimation errors, etc., might cause the robot to make/break contact at a different contact position and/or timing than the planned trajectory. This often leads to constraint violations in the form of slipping or collision of the robot with its environment causing safety hazards. 
\begin{figure}[!t]
\centering
\includegraphics[scale=0.4]{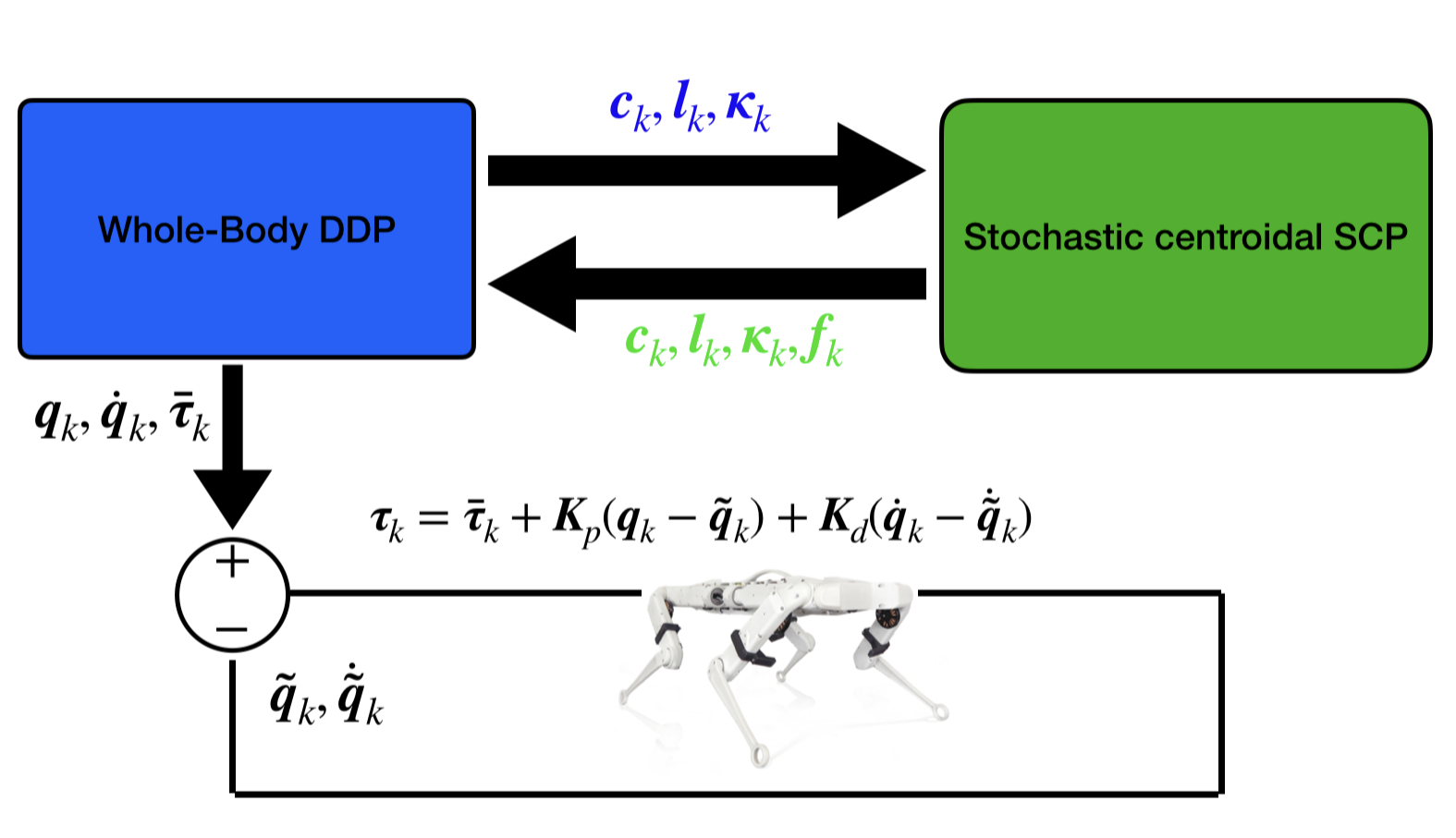}
\caption{Robust trajectory optimization framework alternating between centroidal states of whole-body DDP motions and stochastic centroidal SCP motions.}
\label{fig: framework}
\end{figure}    
Most of the work that took into account uncertainties in the control of legged locomotion mainly resorts to linear models, such as the Linear Inverted Pendulum Model (LIPM). For instance, Villa et al.~\cite{villa2017} used a tube-based linear Robust Model Predictive Control (RMPC) to plan Center of Mass (CoM) trajectories subject to worst-case disturbances on the Center of Pressure (CoP). To reduce the conservativeness of RMPC, Gazar et al.~\cite{gazar2020} resorted to linear stochastic MPC subject to stochastic additive disturbances on the dynamics and CoM linear chance-constraints. Yeganegi et al.~\cite{yeganegi2019} used Bayesian Optimization to learn cost function weights to achieve robust walking motions under different uncertainties. However, assuming fixed height for the CoM and zero angular momentum around the CoM limits the range of motions and cannot plan trajectories in non co-planar multi-contact scenarios. 

There has been some effort in the literature to take into account  uncertainties in the nonlinear trajectory optimization problem for legged locomotion. In these approaches, Differential Dynamic Programming (DDP) has been the method of choice as it takes into account the structure of the underlying optimal control problem to find the optimal trajectories and control policy \cite{tassa2012synthesis,mastalli2020crocoddyl}. In an effort to take uncertainties into account, Mordatch et al.~\cite{mordatch2015} used an offline ensemble DDP enabling them to transfer the whole-body trajectories to a real humanoid robot. Recently, Hammoud et al.~\cite{hammoud2021} used a risk-sensitive version of  DDP to optimize for impedance gains to deal with uncertainties in contact. Despite the appealing convergence properties of DDP-based approaches, inequality constraints can only be considered implicitly in the form of penalties, which makes it difficult to reason about the effect of uncertainties on the constraint satisfaction robustness. Tangent to that work, Darnach et al. ~\cite{drnach2021} used Expected Residual Minimization (ERM) to solve a contact-implicit stochastic complementarity problem. However, contact-implicit trajectory optimization can be quite hard to tune, computationally demanding, and gets easily stuck in local minimas. A middle-ground model between whole body and template models for legged locomotion is the centroidal momentum dynamics~\cite{orin2008}. While this model captures the dynamics between the external forces and the centroidal states exactly~\cite{wieber2016}, it is relatively low dimensional and the nonlinearity structure in the model is well understood \cite{herzog2016structured}. Hence, it is more tractable to use direct methods with explicit consideration of the force and CoM constraints and design force trajectories that are robust to the different types of uncertainties in the model. In this work, we propose to solve stochastic trajectory optimization problems for centroidal momentum generation of legged robots in the presence of additive uncertainties in the model dynamics and uncertainties in the contact location. We write down the friction cone constraints as chance constraints and study the effect of the uncertainties on the satisfaction of these constraints. 
Our work is related to~\cite{brasseur2015robust,dai2016planning}, where the motion of the CoM was restricted within certain bounds to make the underlying optimal control problem convex. In our work instead, we take into account different types of realistic uncertainties in the trajectory optimization problem to find trajectories that are robust against them and are better transferable to the real world. Moreover, compared to~\cite{brasseur2015robust,dai2016planning}, we construct the problem as a stochastic optimization rather than Min-max formulation which is a less conservative formulation and facilitates the generation of more complex behaviours. 

Our main contributions are: 1) We propose to use stochastic trajectory optimization using Sequential Convex Programming (SCP) for generating robust centroidal momentum trajectories subject to additive uncertainties in the dynamics, as well as parametric contact position uncertainties. 2) By considering chance constraints on the friction cones, this is the first work (to the best of our knowledge) that designs controllers for legged robots that generates robust force trajectories subject to contact location uncertainty.
In the same spirit as~\cite{herzog2016structured}, we propose a whole-body trajectory planning framework that alternates between whole-body motion planning and centroidal momentum planning. Contrary to~\cite{herzog2016structured}---but similarly to \cite{budhiraja2019dynamics}---we use whole-body DDP instead of a kinematics optimization, and use the solution to warm-start the stochastic SCP centroidal momentum solver. The resulting (robust) momentum trajectories are later used by the whole-body DDP again to track the resulting robust centroidal trajectories and contact force trajectories. This way, we make sure that the generated momentum trajectories are robust while being consistent with the robot whole-body dynamics. 3) Finally, we run 400 Monte-carlo simulations on the open-source quadruped robot Solo \cite{grimminger2020open} in a Pybullet simulation environment \cite{coumans2016pybullet} for dynamic trotting and bounding gaits while applying different disturbances. We show that Stochastic trajectory optimization is able to complete all the motions safely while reducing feet slippage, and achieving better centroidal tracking performance over the deterministic planning \footnote{\href{https://www.youtube.com/watch?v=IZPyJZe7UDc}{submission video}.}.
\section{Background}
\label{sec:background}
\subsection{Notation}
\label{subsec:notation}
A random variable $x$ following a distribution $\mathcal{Q}$ is denoted as $x \sim \mathcal{Q}$, with $\mathbb{E}[x]$ being the expected value of $x$, and $\boldsymbol{\Sigma}_x \triangleq \mathbb{E}[(\boldsymbol{x}-\mathbb{E}[\boldsymbol{x}])(\boldsymbol{x}-\mathbb{E}[\boldsymbol{x}])^\top]$. The weighted $l_2$ norm is denoted as $\norm{\boldsymbol{y}}_{\boldsymbol{P}} \triangleq \boldsymbol{y}^\top \boldsymbol{P} \boldsymbol{y}$.
\subsection{Robot Dynamics}
The full-body dynamics of a floating-base robot in contact with the environment can be derived using Euler-Lagrange equations of motion as follows~\cite{wieber2016}:
\begin{eqnarray}
\label{eq:manipulator dynamics}
\boldsymbol{M}(\boldsymbol{q})\ddot{\boldsymbol{q}} + \boldsymbol{h}(\boldsymbol{q}, \dot{\boldsymbol{q}}) = \sum^{n_c}_{i=0} \boldsymbol{J}^\top_{e,i}(\boldsymbol{q})\boldsymbol{\lambda}_{e,i} +\boldsymbol{S}^\top \boldsymbol{\tau}_q , 
\end{eqnarray}
where $\boldsymbol{q} = \begin{bmatrix}
\hat{\boldsymbol{x}}^\top, \hat{\boldsymbol{q}}^\top, \boldsymbol{\theta}^\top
\end{bmatrix}^\top \in  \mathbb{R}^3 \times \mathbb{SO}(3) \times \mathbb{R}^n$ represents the generalized robot position characterizing the robot's floating base pose (position and orientation) w.r.t. an inertial frame $\mathcal{I}$, and the joint positions respectively. $\boldsymbol{M}(\boldsymbol{q}) \in \mathbb{R}^{(n+6) \times (n+6)}$ denotes the inertia matrix, and $\boldsymbol{h}(\boldsymbol{q}, \dot{\boldsymbol{q}}) \in \mathbb{R}^{n+6}$ is the vector capturing the Coriolis, centrifugal, gravity and joint friction forces. $\boldsymbol{J}_{e,i}$ is the associated jacobian of the $i$-th end-effector wrench $\boldsymbol{\lambda}_{e,i}$ acting on the environment. Finally, $\boldsymbol{S} = \begin{bmatrix}\boldsymbol{0}_{(n \times 6)},\, \boldsymbol{I}_n \end{bmatrix}$ is the selector matrix of the actuated joint torques $\boldsymbol{\tau}_q$. We can further split (\ref{eq:manipulator dynamics}) into its under-actuated and actuated parts;
\begin{subequations}
\begin{align}
\boldsymbol{M}^u(\boldsymbol{q})\ddot{\boldsymbol{q}} + \boldsymbol{h}^u(\boldsymbol{q}, \dot{\boldsymbol{q}}) &= \sum^{n_c}_{i=0} \boldsymbol{J}^{u^\top}_{e,i}(\boldsymbol{q})\boldsymbol{\lambda}_{e,i}, \label{eq:under-actuated dynamics}\\
\boldsymbol{M}^a(\boldsymbol{q})\ddot{\boldsymbol{q}} + \boldsymbol{h}^a(\boldsymbol{q}, \dot{\boldsymbol{q}}) &=  \sum^{n_c}_{i=0} \boldsymbol{J}^{a^\top}_{e,i}(\boldsymbol{q})\boldsymbol{\lambda}_{e,i} + \boldsymbol{\tau}_q. 
\end{align}
\end{subequations}
By writing down the floating-base dynamics for the CoM instead of the floating base position $\hat{x}$,
we obtain the following relationship between the centroidal momentum dynamics $\dot{\boldsymbol{h}}_\mathcal{G} $ and the generalized velocities $\dot{\boldsymbol{q}}$
\begin{equation}
\label{eq:centroidal momentum matrix}
 \dot{\boldsymbol{h}}_\mathcal{G} = \begin{bmatrix}\dot{\boldsymbol{\kappa}} \\ \dot{\boldsymbol{l}}\end{bmatrix} =  \dot{\boldsymbol{A}}_\mathcal{G}(\boldsymbol{q}) \ddot{\boldsymbol{q}} + \dot{\boldsymbol{A}}_\mathcal{G} (\boldsymbol{q}) \dot{\boldsymbol{q}},
\end{equation}
via the \textit{Centroidal Momentum Matrix} (CMM) \mbox{$\boldsymbol{A}_\mathcal{G} \in \mathbb{R}^{6 \times (n+6)}$}~\cite{orin2008}. The angular and linear momenta are denoted as $\boldsymbol{\kappa}$ and $\boldsymbol{l} \in \mathbb{R}^3$ respectively. Given (\ref{eq:centroidal momentum matrix}),  we are interested in planning desired centroidal momentum trajectories that satisfy the following Newton-Euler dynamics:
\begin{equation}
\label{eq:centroidal momentum dynamics}
\dot{\boldsymbol{h}} = \begin{bmatrix}\sum^{n_c}_{i=0}(\boldsymbol{p}_{e,i} + \boldsymbol{R}_{e,i}^{x,y}\boldsymbol{\zeta}_{e,i} - \boldsymbol{c}) \times \boldsymbol{f}_{e,i} + \boldsymbol{R}^z_{e,i}\tau_{e,i} \\ m \boldsymbol{g} + \sum^{n_c}_{i=0} \boldsymbol{f}_{e,i} \end{bmatrix}
\end{equation}
where $\boldsymbol{c} \in \mathbb{R}^3$  represents the robot's CoM, at which the total mass $m$ of the robot is concentrated. $\boldsymbol{p}_{e,i} \in \mathbb{R}^3$ is the $i$-th end-effector's contact position, with $\boldsymbol{\zeta}_{e,i} \in \mathbb{R}^2$ being the local $\textit{Center of Pressure}$ (CoP). $\boldsymbol{f}_{e,i} \in \mathbb{R}^3$ and \mbox{$\tau_{e,i} \in \mathbb{R}$} represent end-effector's contact forces and torque for flat-footed robots, respectively. The rotation matrix $\boldsymbol{R}_{e,i} \in \mathbb{SO}(3)$ captures the contact normals mapping quantities from the $i$-th end-effector's frame to the inertial frame. Note that for point-footed robots, which we consider from now on, $\boldsymbol{\zeta}_{e, i}$ and $\tau_{e,i}$ are always null, but the same analysis still holds for flat-footed robots.
\subsection{Centroidal Momentum Trajectory Optimization}
\label{sub-section:nominal NOCP}
\newtheorem{Problem}{Problem}
First we present the deterministic nonlinear discrete-time optimal control problem (OCP) for centroidal momentum trajectory optimization with fixed contact position and timing.
\begin{Problem} Nominal Optimal Control Problem (NOCP)
\label{NOCP}
\begin{subequations}
\begin{align}
\label{eq: centoridal OCP} 
&\min_{\substack{\boldsymbol{x}, \boldsymbol{u}}}. \,\,\, l_f(\boldsymbol{x}_N) + \sum^{N-1}_{i=0}l(\boldsymbol{x}_k, \boldsymbol{u}_k) \\
&\text{s.t.} \nonumber \\
& \quad\begin{bmatrix} 
    \boldsymbol{c}_{k+1} \\ \\
	\boldsymbol{l}_{k+1} \\ \\ \boldsymbol{\kappa}_{k+1}
	\end{bmatrix} = \begin{bmatrix}
	\boldsymbol{c}_{k} + \frac{1}{m}\boldsymbol{l}_k\Delta_k \\ \\
	\boldsymbol{l}_{k} + m \boldsymbol{g}\Delta_k + \sum^{n_c}_{i=0} \boldsymbol{f}_{{e,i}_k} \Delta_k\\ \\
	\boldsymbol{\kappa}_{k} + \sum^{n_c}_{i=0} (\boldsymbol{p}_{{e,i}_k} - \boldsymbol{c}_k) \times \boldsymbol{f}_{{e,i}_k} \Delta_k  
	\end{bmatrix}, \label{eq:discrete centroidal dynamics}\\
 &\quad-\mu \mathfrak{f}^z_{{e,i}_k} \leq\mathfrak{f}^{x}_{{e,i}_k} \leq  \mu \mathfrak{f}^z_{{e,i}_k}, \quad \mathfrak{f}^z_{{e,i}_k} \geq 0, \label{eq:friction pyramid constraints x-direction}\\
 &\quad-\mu \mathfrak{f}^z_{{e,i}_k} \leq\mathfrak{f}^{y}_{{e,i}_k} \leq  \mu \mathfrak{f}^z_{{e,i}_k},\quad\mathfrak{f}^z_{{e,i}_k} \geq 0, \label{eq:friction pyramid constraints y-direction}\\
&\quad|\boldsymbol{p}_{{e,i}_k} - \boldsymbol{c}_k| \leq \mathcal{L}^{\max}_{e,i},      \label{eq:reachability constraints} \\
&\quad\boldsymbol{x}_0 = \boldsymbol{x}(0),  \label{eq:initial constraints}\\
&\quad\boldsymbol{x}_f = \boldsymbol{x}(N),  \label{eq:terminal constraints}\\
&\quad\forall k \in \{0,1,..,N-1\}, 
\end{align}
\end{subequations}
\end{Problem}
where $\boldsymbol{x}=(\boldsymbol{x}_0, \dots, \boldsymbol{x}_{N})$ with $\boldsymbol{x}_k \in \mathbb{R}^9 = (\boldsymbol{c}_k, \boldsymbol{l}_k, \boldsymbol{\kappa}_k )$, and $\boldsymbol{u}= (\boldsymbol{u}_0, \dots, \boldsymbol{u}_{N-1})$ with $\boldsymbol{u}_k \in \mathbb{R}^{3n_c} = (\boldsymbol{f}_{e,0}, \dots, \boldsymbol{f}_{e, n_c})$ are the states and control optimizers along the control horizon N. The centroidal momentum dynamics are discretized with a time-step $\Delta_k$ using an explicit Euler integration scheme \eqref{eq:discrete centroidal dynamics}, where \eqref{eq:initial constraints}-\eqref{eq:terminal constraints} represents the initial and final conditions respectively. In order to avoid contact slippage, the local contact forces in the end-effector frame ($\boldsymbol{\mathfrak{f}} = \boldsymbol{R}^T \boldsymbol{f} $) are constrained inside the linearized friction cone constraints \eqref{eq:friction pyramid constraints x-direction}-\eqref{eq:friction pyramid constraints y-direction}, where the static coefficient of friction is denoted as $\mu$ with the vertical component of the force being positive. Finally, the CoM is constrained to be within the leg length reachability limits \eqref{eq:reachability constraints}. 
\newtheorem{assumption}{Assumption}
\section{Stochastic Optimal Control for Centroidal Momentum Trajectory Optimization}
In this section we present a stochastic version of problem ($\ref{NOCP}$) that takes into account additive stochastic uncertainties on the centroidal momentum dynamics as well as  contact position uncertainties subject to friction pyramid chance constraints. We consider the following discrete-time stochastic nonlinear OCP:
\begin{Problem}
\label{SOCP}
Stochastic Optimal Control Problem (SOCP)
\begin{subequations}
\begin{align}
	&\min_{\substack{\boldsymbol{x}, \boldsymbol{u}}}. \,\,\, l_f(\boldsymbol{x}_N) + \sum^{N-1}_{i=0}l(\boldsymbol{x}_k, \boldsymbol{u}_k) \\
	&\text{s.t.} \nonumber \\
	&\quad \boldsymbol{x}_{k+1} = \boldsymbol{f}(\boldsymbol{x}_k, \boldsymbol{u}_k, \boldsymbol{\theta}_k, \boldsymbol{w}_k), \label{eq:discrete stochastic centroidal dynamics}\\
	&\quad \text{Pr}(\boldsymbol{H}\boldsymbol{x_k} \leq \boldsymbol{h}) \geq \alpha_x, \label{eq:state chance constraints}\\
	&\quad \text{Pr}(\boldsymbol{G}\boldsymbol{u}_k \leq \boldsymbol{g}) \geq \alpha_u, \label{eq:control chance constraints}\\
	&\quad \boldsymbol{x}_0 = \boldsymbol{x}(0),  \label{eq:initial constraints_stoch}\\
	&\quad \boldsymbol{x}_f = \boldsymbol{x}(N),  \label{eq:terminal constraints_stoch}\\
	&\quad  \forall k \in \{0,1,..,N-1\}. 
\end{align}
\end{subequations}
\end{Problem}
With an abuse of notation from Problem \eqref{NOCP}, $\boldsymbol{x}_k$ and $\boldsymbol{u}_k$ will be considered the stochastic state and control policies evolving according to the parametric and additive stochastic disturbance realizations $\boldsymbol{\theta}_k$, and $\boldsymbol{w}_k$. \eqref{eq:state chance constraints}-\eqref{eq:control chance constraints} are the state and control polytopic joint chance constraints with $\alpha_x$ and $\alpha_u$ being the probability levels of state and control constraint satisfaction respectively.
\begin{assumption}(i.i.d. Gaussian disturbances) \\
\label{eq:gaussian noise assumption}
$\boldsymbol{\theta}_k\sim\mathcal{N}(\mathbb{E}[\boldsymbol{\theta}_k],  \boldsymbol{\Sigma}_{\boldsymbol{\theta}_k})$, and  $\boldsymbol{w}_k \sim \mathcal{N}(\mathbb{E}[\boldsymbol{w}_k], \boldsymbol{\Sigma}_{\boldsymbol{w}_k})$ are assumed to be independent and identically distributed  (i.i.d.) disturbance realisations following Gaussian distributions. $\mathbb{E}[\boldsymbol{\theta}_k] = \boldsymbol{p}_{{e,i}_k}$, and $\boldsymbol{\Sigma}_{\boldsymbol{\theta}_k} \in \mathbb{R}^{(3n_c \times 3n_c)}$ represent the mean and covariance of the contact positions respectively. $\mathbb{E}[\boldsymbol{w}_k] = \boldsymbol{0}$ and $\boldsymbol{\Sigma}_{\boldsymbol{w}_k} \in \mathbb{R}^{9 \times 9}$ are the mean and covariance of the additive noise on the centroidal dynamics.  
\end{assumption} 
\subsection{Individual Chance Constraints Reformulation}
Solving the above joint chance constraints \eqref{eq:state chance constraints}-\eqref{eq:control chance constraints} involves the  integration of multi-dimensional Gaussian Probability Density Functions (PDFs), which becomes computationally intractable for high dimensions. One effective solution is to use Boole's inequality:
\begin{equation}
\label{eq:boole's inequality}
\text{Pr}(\bigvee^n_{i=1} \boldsymbol{C}_i)\leq \sum^n_i \text{Pr} (\boldsymbol{C}_i) 
\end{equation} 
 as a conservative union bound on the joint chance constraints \cite{Ono2010}. We can rewrite the complement of the state chance constraints as a conjunction of individual chance constraints as follows:
\begin{align}
\label{eq:disjunctive state chance constraints}
(\ref{eq:state chance constraints}) &=  \text{Pr}(\bigwedge^{l_x}_{i=1} \boldsymbol{H}_i \boldsymbol{x} \leq h_i) \geq \alpha_x, \nonumber 
\end{align} 
which can be written conservatively as 
\begin{align}
\text{Pr}(\bigvee^{l_x}_{i=1}\boldsymbol{H}_i \boldsymbol{x} > h_i) \leq 1-\alpha_{x}.
\end{align}    

By applying Boole's inequality on the above equation, and allocating constraint violation risk equally $\epsilon_{x_i} = (1-\alpha_x)/l_x$, with $l_x$ being the number of intersecting hyper-planes forming the state joint polytopic constraint, we reach
\begin{align}
\label{eq:indvidual state chance constraints}
(\ref{eq:disjunctive state chance constraints})\xLeftarrow[]{(\ref{eq:boole's inequality})} & \sum^{l_x}_{i=1} \text{Pr}(\boldsymbol{H}_i \boldsymbol{x} > h_i) \leq \epsilon_{x_i} \nonumber \\  \equiv & \sum^{l_x}_{i=1}\text{Pr}(\boldsymbol{H}_i \boldsymbol{x} \leq h_i) \geq 1-\epsilon_{x_i}.
\end{align}
Similarly, control joint chance constraint (\ref{eq:control chance constraints}) can be reformulated as a set of individual chance constraints following the same arguments as before
\begin{equation}
\label{eq:indvidual control chance constraints}
\sum^{l_u}_{i=1}\text{Pr}(\boldsymbol{G}_i \boldsymbol{u} \leq g_i) \geq 1-\epsilon_{u_i}, 
\end{equation}
where $\epsilon_{u_i} = (1-\alpha_u)/l_u$ is the equally distributed control constraint risk. 
\begin{remark}
Allocating risk of constraint violations equally can be quite conservative since one would preferably allocate more risk to active constraints over inactive ones. Another approach can be optimizing for allowable violation for each constraint as in \cite{Ma2012}, which involves higher computational complexity.  
\end{remark}
\subsection{Deterministic Reformulation of Individual Chance Constraints }
Solving the chance constraints (\ref{eq:indvidual state chance constraints})-(\ref{eq:indvidual control chance constraints}), requires propagating the uncertainty through the nonlinear dynamics. We adopt a linearization-based covariance propagation as in \cite{zhu2019}\cite{lew2020}. Using a state-feedback control policy $\boldsymbol{u}_k = \boldsymbol{v}_k + \boldsymbol{K}_k(\boldsymbol{x}_k - \boldsymbol{s}_k)$, where $\boldsymbol{K}_k$ are pre-stabilizing feedback gains, then the mean and covariance of the dynamics evolve as  
\begin{subequations}
\begin{align}
 \boldsymbol{s}_{k+1} & \approx \bar{\boldsymbol{ f}}(\boldsymbol{s}_k, \boldsymbol{v}_k, \boldsymbol{p}_{e,k}, \boldsymbol{0})+ \boldsymbol{A}_k (\boldsymbol{s}_k-\boldsymbol{s}^j_k) + \boldsymbol{B}_k(\boldsymbol{v}_k-\boldsymbol{v}^j_k), \label{eq:linearized dynamics}\\
 \boldsymbol{\Sigma}_{\boldsymbol{x}_{k+1}} &=  \boldsymbol{A}_{\text{cl}}\boldsymbol{\Sigma}_{\boldsymbol{x}_k}  \boldsymbol{A}_{\text{cl}}^\top + \boldsymbol{C}_k\boldsymbol{\Sigma}_{\boldsymbol{\theta}}\boldsymbol{C}^\top_k + \boldsymbol{\Sigma_w}, \label{eq:covariance propagation}
\end{align}
\end{subequations}
where $\bar{\boldsymbol{f}}$ is the nominal nonlinear dynamics estimated at current mean of the state $\boldsymbol{s}^j_k$ and controls $\boldsymbol{v}^j_k$ of the $jth$ trajectory. $\boldsymbol{\Sigma}_{\boldsymbol{x}_0} = \boldsymbol{0}$, and $\boldsymbol{A}_\text{cl} \triangleq \boldsymbol{A}_k + \boldsymbol{B}_k \boldsymbol{K}_k$ is the closed-loop dynamics. $\boldsymbol{A}_k \triangleq \frac{\partial}{\partial \boldsymbol{s}} \boldsymbol{f}(\boldsymbol{s}_k, \boldsymbol{v}_k, \boldsymbol{p}_{e,k}, \boldsymbol{0})\vert_{(\boldsymbol{s}^j_k, \boldsymbol{v}^j_k)}$ is the Jacobian of the dynamics w.r.t. the state. $\boldsymbol{B}_k \triangleq  \frac{\partial}{\partial \boldsymbol{v}} \boldsymbol{f}(\boldsymbol{s}_k, \boldsymbol{v}_k, \boldsymbol{p}_{e,k}, \boldsymbol{0})\vert_{(\boldsymbol{s}^j_k, \boldsymbol{v}^j_k)}$ is the Jacobian of the dynamics w.r.t. controls. Finally, $\boldsymbol{C}_k \triangleq \frac{\partial}{\partial \boldsymbol{p}_{e,k}} \boldsymbol{ f}(\boldsymbol{s}_k, \boldsymbol{v}_k, \boldsymbol{p}_{e,k}, \boldsymbol{0})\vert_{(\boldsymbol{s}^j_k, \boldsymbol{v}^j_k)}$ represents the Jacobian of the dynamics w.r.t. the contact positions. 
\begin{remark} 
Other approaches can be used for uncertainty propagation through nonlinear dynamics like unscented-based transforms \cite{plancher2017}, or Generalized Polynommial Chaos (gPC) \cite{nakka2019}. These methods can lead to more accurate estimate of the propagated uncertainty at the cost of significant increase in complexity. Since computational efficiency is more important in our case (especially for online re-planning of the trajectories), we prefer to not use these methods.
\end{remark}
Based on Assumption (\ref{eq:gaussian noise assumption}) and the covariance propagation in (\ref{eq:covariance propagation}), we seek the least conservative upper bounds on the state and controls individual chance-constraints (\ref{eq:state chance constraints})-(\ref{eq:control chance constraints}). Using the inverse of the \textit{Cumulative Density Function} (CDF) $\phi^{-1}$ of a Gaussian distribution, we arrive to a deterministic reformulation of the chance constraints:
\begin{subequations}
\begin{align}
\label{eq:chance-constraints backoffs}
    \boldsymbol{H}_i \boldsymbol{s}_k \leq h_i - \eta_{x_{i,k}}, \\
    \boldsymbol{G}_i \boldsymbol{v}_k \leq g_i - \eta_{u_{i,k}},
\end{align}
\end{subequations}
where $\eta_{x_{i,k}} = \phi^{-1}(1-\epsilon_{x_i}) \norm{\boldsymbol{H}_i}_{\boldsymbol{\Sigma}_k}$ and  $\eta_{u_{i,k}} = \phi^{-1}(1-\epsilon_{u_i}) \norm{\boldsymbol{G}_i\boldsymbol{K}_k}_{\boldsymbol{\Sigma}_k}$ are known as the state and control back-off bounds ensuring the satisfaction of the individual chance constraints (\ref{eq:state chance constraints})-(\ref{eq:control chance constraints}), respectively.
\subsection{Deterministic Reformulation of SOCP}
Given the previous reformulation of the individual chance constraints, we can write down the following NOCP.
\begin{Problem}
\label{problem:CCOCP}
NOCP with reformulated individual chance-constraints:
\begin{subequations}
\begin{align}
	&\min_{\substack{\boldsymbol{s}, \boldsymbol{v}}}. \,\,\, l_f(\boldsymbol{s}_N) + \sum^{N-1}_{i=0}l(\boldsymbol{s}_k, \boldsymbol{v}_k) \\
	&\text{s.t.} \nonumber \\
	&\quad \boldsymbol{s}_{k+1} = \boldsymbol{f}(\boldsymbol{s}_k, \boldsymbol{v}_k, \boldsymbol{p}_{e,k}, \boldsymbol{0}), \label{eq:nonlinear deterministic dynamics} \\
    &\quad \boldsymbol{H}_{i,k} \boldsymbol{s}_k \leq h_{i,k} - \eta_{x_{i,k}} \quad\quad  \forall i \in \{1, 2,..,l_x\}, \label{eq:detereminstic reformulation of state constraints} \\
    &\quad \boldsymbol{G}_{i,k} \boldsymbol{v}_k \leq g_{i,k} - \eta_{u_{i,k}} \quad\quad\,  \forall i \in \{1, 2,..,l_u\},\label{eq:detereminstic reformulation of control constraints} \\
	&\quad \boldsymbol{s}_0 = \boldsymbol{s}(0), \\
	&\quad \boldsymbol{s}_f = \boldsymbol{s}(N), \\
	&\quad  \forall k \in \{0,1,..,N-1\}.
\end{align}
\end{subequations}
\end{Problem}
where \eqref{eq:nonlinear deterministic dynamics} is now the mean of the nonlinear dynamics. In order to solve the the above nonlinear OCP, we resort to Sequential Convex Programming (SCP), which we explain in the next subsection.
\subsection{SCP with L1 Trust Region  Penalty Cost}
\label{sub-section:SCP}
SCP attempts to solve nonlinear OCPs by successively linearizing the dynamics, costs and constraints to solve a convex sub-problem at every iteration. The dynamics are linearized with a first-order Taylor expansion around the previous state and control trajectories computed at the $j$-th succession. 
Successive linearization introduces two well-known problems \cite{Ma2012}.

1) Artificial infeasibility: the problem becomes infeasible even if the original nonlinear problem is feasible. The  most evident example of this arises when the problem is linearized about an unrealistically short time  horizon, so that there is no feasible control input that can  satisfy the prescribed dynamics and constraints. 2) Artificial unboundedness: the solution takes steps far away from the validity of the linear model.
In order to mitigate artificial unboundedness, a trust-region constraint is employed. Different approaches are adapted to tackle artificial infeasiblity. In \cite{Ma2012}, the authors employ hard constraints and virtual controls as slack variables on the constraints. However, \cite{bonalli2019}
enforced hard constraints on the dynamics and convex soft penalties on the rest of the constraints along with trust region constraints. In this work we follow the same rationale as \cite{bonalli2019,schulman2014}, where the trust region constraints $c_i(\boldsymbol{x}) \leq 0$ are enforced as $l_1$ penalty cost in the form of 
\begin{subequations}
\begin{align}
\label{eq:trust region l1 penalty}	
\argmin.\{\argmax.\gamma(c_i(\boldsymbol{x}), 0)\}, \\
c_i(\boldsymbol{x}) = |\boldsymbol{x}_k-\boldsymbol{x}^j_k| - \Omega.
\end{align}
\end{subequations}
where $\Omega$ is the trust region radius. Notice that the above $l_1$ penalty cost is exact---meaning that as the penalty weight $\gamma$ gets infinitely large, the constraint violations are driven to zero. Even though (\ref{eq:trust region l1 penalty}) is non-differentiable, yet it can be solved efficiently by introducing a slack variable $t$ as follows:
\begin{subequations}
\begin{align}
	\label{eq:dual trust region l1 penalty}	
	&\min_t. \quad \gamma  t \\
	&\text{s.t.} \nonumber \\
	&\quad |\boldsymbol{x}-\boldsymbol{x}^j| - \Omega \leq t, \\
	&\quad - t \leq 0.
\end{align}
\end{subequations}
In order to solve problem (\ref{problem:CCOCP}), we solve a sequence of Quadratic Programs (QPs) in problem (\ref{problem:QP}), accompanied by a trust region update mechanism based on the accuracy ratio of the linearized model w.r.t. the nonlinear model as in \cite{bonalli2019}\cite{lew2020}. 
\begin{Problem}
\label{problem:QP}
Convexified QP at the $j$-th SCP iteration:
\begin{subequations}
\begin{align}
	&\min_{\substack{\boldsymbol{s}, \boldsymbol{v}, \boldsymbol{t}}}. \,\,\, l_f(\boldsymbol{s}_N) + \sum^{N-1}_{i=0}l(\boldsymbol{s}_k, \boldsymbol{v}_k) + \gamma^j \sum^{N}_{i=0} t_k \\
    &\text{s.t.} \nonumber \\
	&\quad\boldsymbol{s}_{k+1} = \bar{\boldsymbol{f}}(\boldsymbol{s}_k, \boldsymbol{v}_k, \boldsymbol{p}_{e,k}, \boldsymbol{0}) + \boldsymbol{A}_k (\boldsymbol{s}_k-\boldsymbol{s}^j_k) + \boldsymbol{B}_k(\boldsymbol{v}_k-\boldsymbol{v}^j_k), \\
	&\quad\boldsymbol{\Sigma}_{k+1} =  \boldsymbol{A}_{\text{cl}}\boldsymbol{\Sigma}_{\boldsymbol{x}_k} \boldsymbol{A}_{\text{cl}}^\top + \boldsymbol{C}_k\boldsymbol{\Sigma}_{\boldsymbol{\theta}}\boldsymbol{C}^\top_k + \boldsymbol{\Sigma_w}, \\
	&\quad\boldsymbol{\Sigma}_0 = \boldsymbol{0}_{9 \times 9}, 
\end{align}
\begin{align}
	  &\quad\boldsymbol{H}_{i, k} \boldsymbol{s}_k \leq h_{i,k} - \phi^{-1}(1-\epsilon_{x_i})\Big(\norm{\boldsymbol{H}_{i,k}}_{\boldsymbol{\Sigma}_k} + \frac{\partial}{\partial{\boldsymbol{z}}}\norm{\boldsymbol{H}_{i,k}}_{\boldsymbol{\Sigma}_k}(\boldsymbol{z}_k-\boldsymbol{z}_k^j)\Big), \nonumber\\ 
    & \hspace{8.5 cm}\forall i \in \{1, 2,..,l_x\}, \label{eq:linearized state chance constraints} \\
    &\quad\boldsymbol{G}_{i, k}\boldsymbol{v}_k \leq g_{i,k} - \phi^{-1}(1-\epsilon_{u_i})\Big(\norm{\boldsymbol{G}_{i, k} \boldsymbol{K}_k}_{\boldsymbol{\Sigma}_k} + \frac{\partial}{\partial{\boldsymbol{z}}}\norm{\boldsymbol{G}_{i, k} \boldsymbol{K}_k}_{\boldsymbol{\Sigma}_k}(\boldsymbol{z}_k-\boldsymbol{z}_k^j)\Big), \nonumber \\
    & \hspace{8.5 cm}\forall i \in \{1, 2,..,l_u\},\label{eq:linearized control chance constraints}\\
	&\quad|\boldsymbol{\kappa}_k-\boldsymbol{\kappa}_k^j| - \Omega^j \leq t_k, \quad -t_k \leq 0, \label{eq:angular momentum trust region constraint}\\
	&\quad\boldsymbol{s}_0 = \boldsymbol{s}(0), \\
	&\quad\boldsymbol{s}_f = \boldsymbol{s}(N), \\
	&\quad\forall k \in \{0,1,..,N-1\}. 
\end{align}
\end{subequations}
\end{Problem}

$\boldsymbol{z}_k \in \mathbb{R}^{9+3n_c} = (\boldsymbol{s}_k,\boldsymbol{v}_k)$ is the concatenated vector of states and controls at time $k$. Constraints (\ref{eq:linearized state chance constraints})-(\ref{eq:linearized control chance constraints}) are the linearized state and control chance constraints, where
\begin{align}
\frac{\partial}{\partial{\boldsymbol{z}}}\norm{\boldsymbol{H}_{i,k}}_{\boldsymbol{\Sigma}_k} = \frac{1}{2\norm{\boldsymbol{H}_{i,k}}_{\boldsymbol{\Sigma}_k}}\Big(2\boldsymbol{H}^\top_i \boldsymbol{\Sigma}_k \frac{\partial}{\partial \boldsymbol{z}}\boldsymbol{H}_{i,k} + \sum^n_{i=0} \sum^n_{j=0} h_i h_j \frac{\partial}{\partial \boldsymbol{z}} \Sigma_{ij}\Big). 
\end{align}
$\frac{\partial}{\partial \boldsymbol{z}} \boldsymbol{\Sigma} \in \mathbb{R}^{9 \times 9 \times (9 + 3n_c)}$ represents the covariance derivative w.r.t. $\boldsymbol{z}$. Notice that this term is more involved since it includes the propagation of the tensor derivatives of the covariance matrix given the current states and controls as well as the previous states and controls as follows:
\begin{equation}
\frac{\partial}{\partial \boldsymbol{z}}\boldsymbol{\Sigma}_{k+1} = \sum^{k-1}_{i=0} \boldsymbol{A}_k \frac{\partial}{\partial \boldsymbol{z}}\boldsymbol{\Sigma}_{k+1|i} \boldsymbol{A}^\top_k + \frac{\partial}{\partial \boldsymbol{z}}\boldsymbol{\Sigma}_{k+1|k}.
\end{equation}
We resort to the autodiff library JAX \cite{jax2018} for such computation. Finally, the trust region constraints (\ref{eq:angular momentum trust region constraint}) are enforced only on the angular momentum $\boldsymbol{\kappa}_k$ since it's the only nonlinear part in the centroidal dynamics. 
\section{Simulations Results}
\label{sec:results}
In this section, we report simulation results for the quadruped robot Solo in the Pybullet simulation environment \cite{coumans2016pybullet}. We compare trajectories generated using centroidal stochastic trajectory optimization against nominal trajectory optimization for trotting and bounding gaits on challenging unknown cluttered terrains. Offline, we warm start the centroidal SCP solver using centroidal trajectories coming from the whole-body DDP solver Croccodyl \cite{mastalli2020crocoddyl}. Then, we optimize whole-body trajectories to track back the optimized centroidal and force trajectories from the SCP solver as illustrated pictorially in Fig.~\ref{fig: framework}. The cost weights for both whole-body DDP and centroidal SCP are summarized in Table~\ref{table:DDP parameters} and Table~\ref{table:scp parameters}, respectively. Both DDP and SCP solvers were discretized with a sampling time of $\Delta_k = 10$ ms for a planning horizon length of $N = 165$, and motion plans were designed on a flat ground with a floor static coefficient of friction $\mu = 0.5$ for both solvers.
\begin{figure}[!t]%
    \begin{minipage}[b]{.1\columnwidth}
    \centering
    \includegraphics[trim=400 100 150 250 ,clip,scale=0.09]{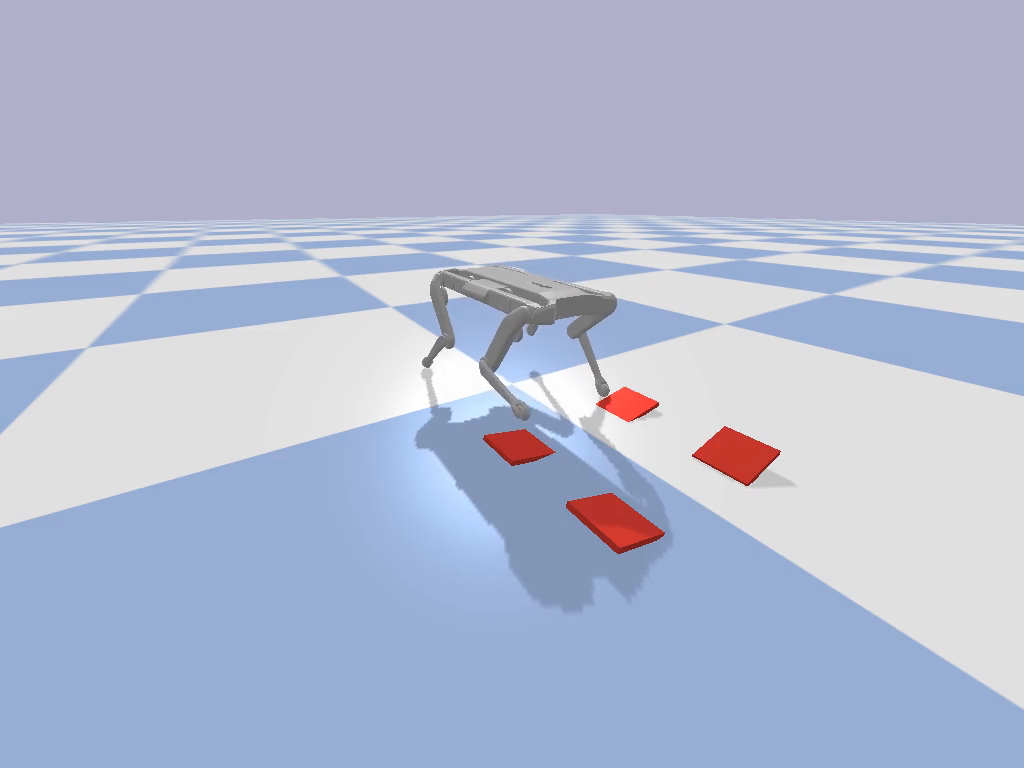}
    \end{minipage}
    \,\,\,
    \begin{minipage}[b]{.1\columnwidth}
    \includegraphics[trim=400 100 150 250 ,clip,scale=0.09]{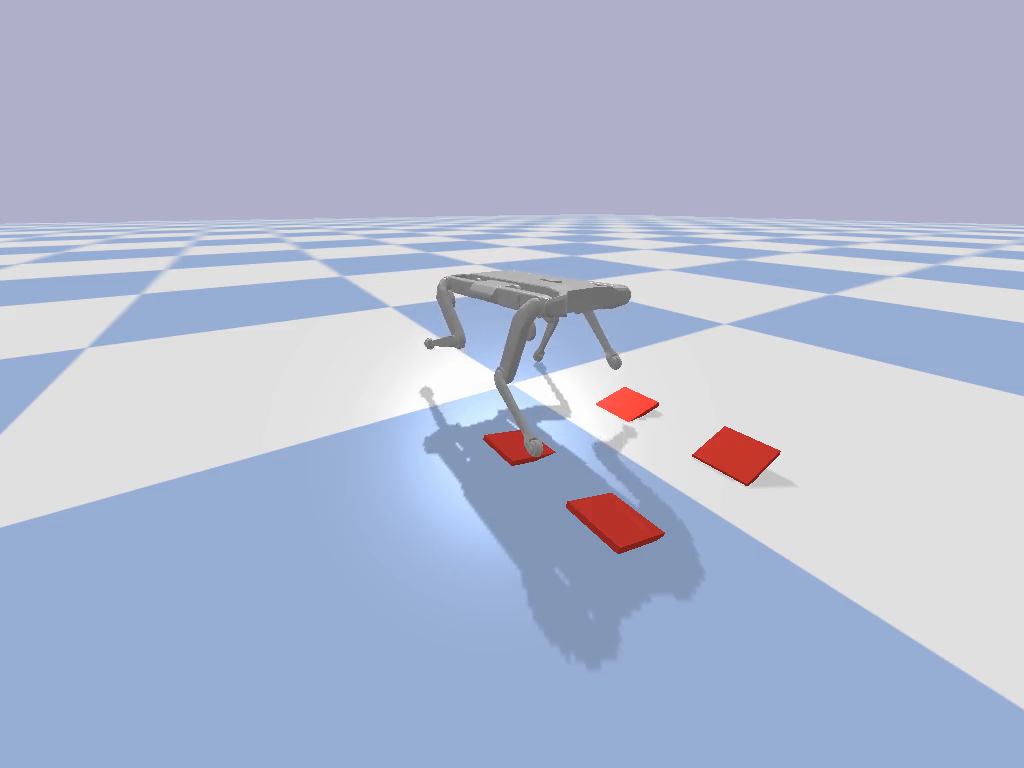}
    \end{minipage}
    \,\,\,
    \begin{minipage}[b]{.1\columnwidth}
    \centering
    \includegraphics[trim=400 100 150 250 ,clip,scale=0.09]{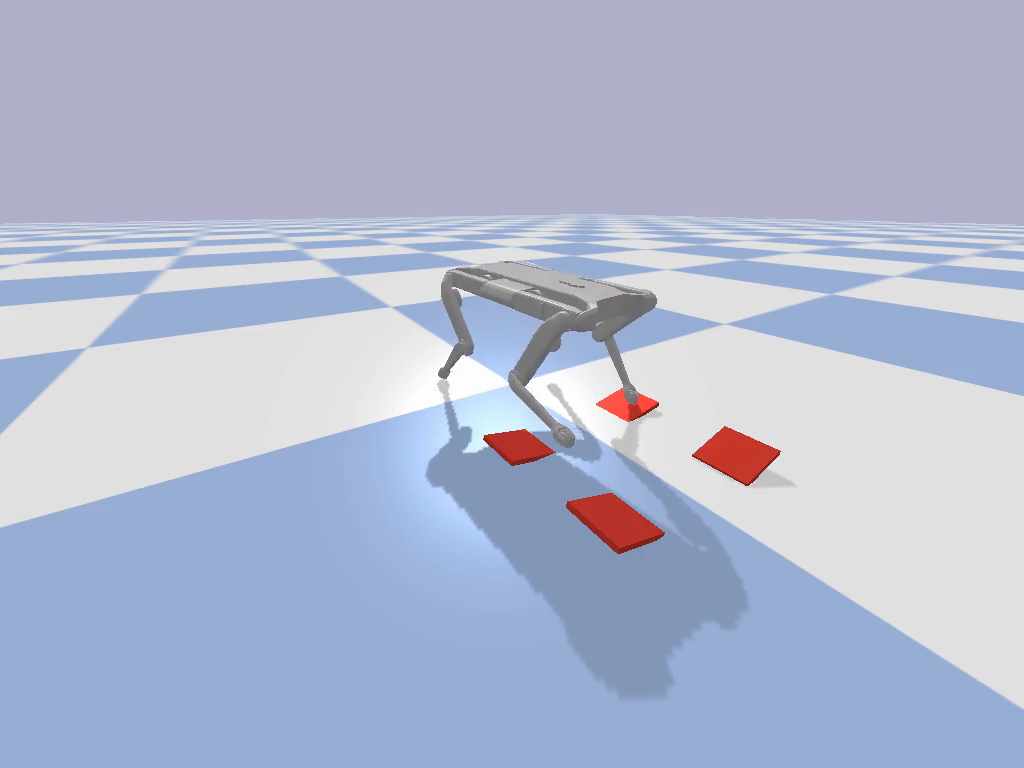}
    \end{minipage}
    \,\,\,
    \begin{minipage}[b]{.1\columnwidth}
    \centering
    \includegraphics[trim=400 100 150 250 ,clip,scale=0.09]{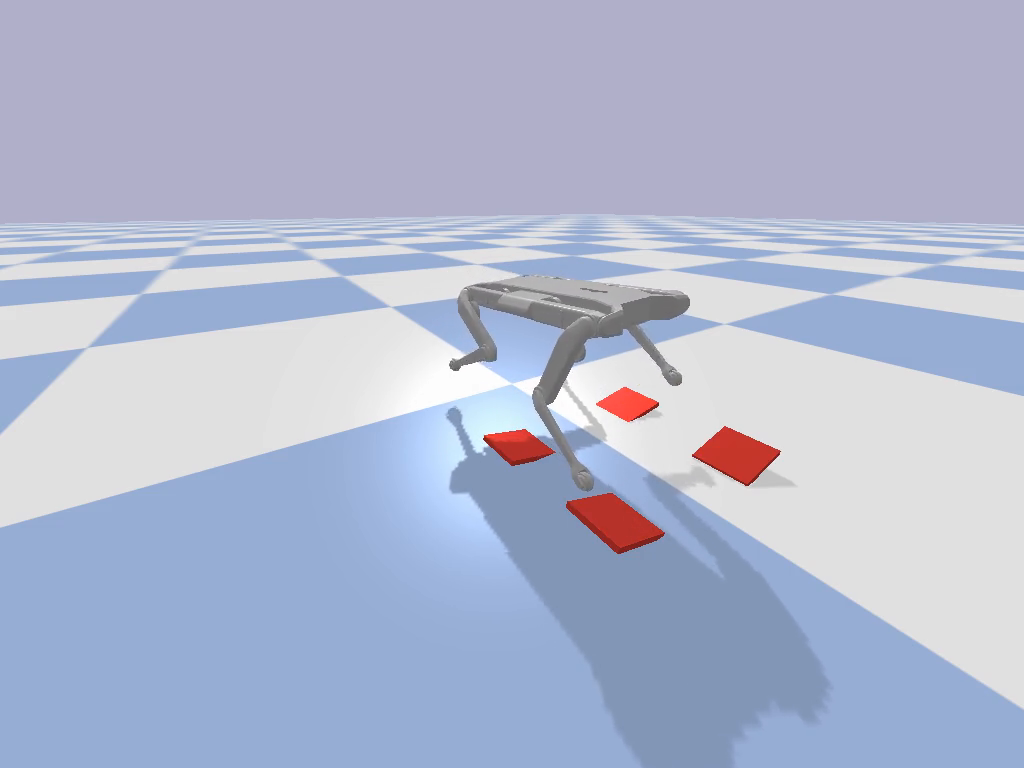}
    \end{minipage}
    \,\,\,
    \begin{minipage}[b]{.1\columnwidth}
    \centering
    \includegraphics[trim=400 100 150 250 ,clip,scale=0.09]{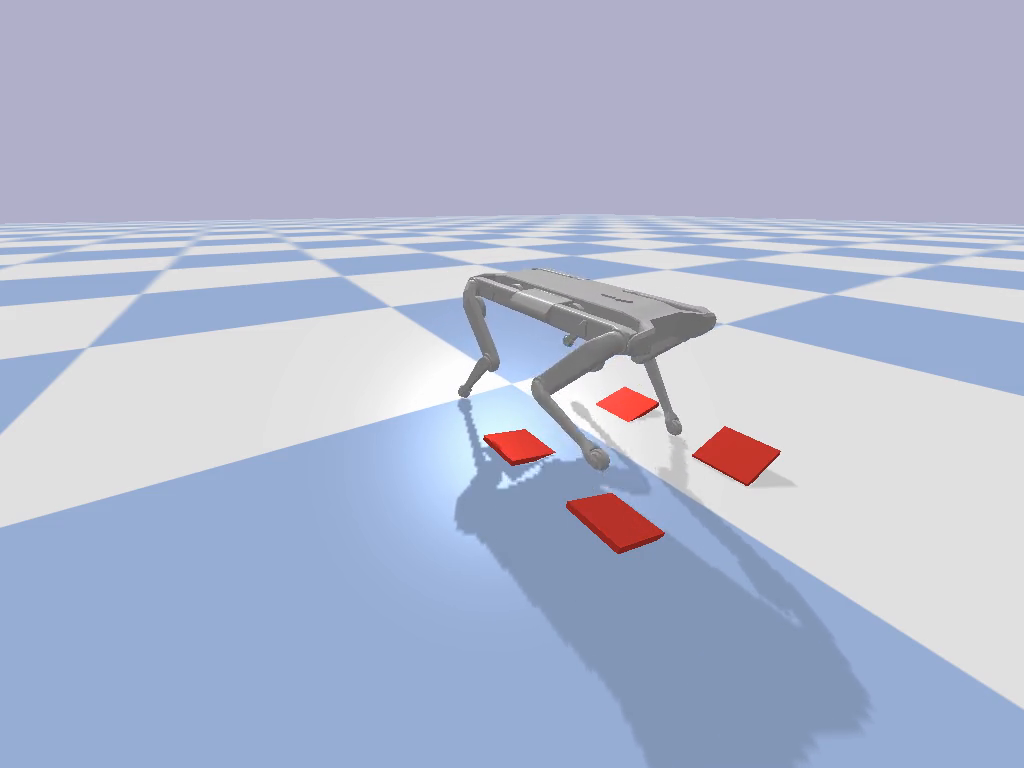}
    \end{minipage}
    \,\,\,
    \begin{minipage}[b]{.1\columnwidth}
    \includegraphics[trim=400 100 150 250 ,clip,scale=0.09]{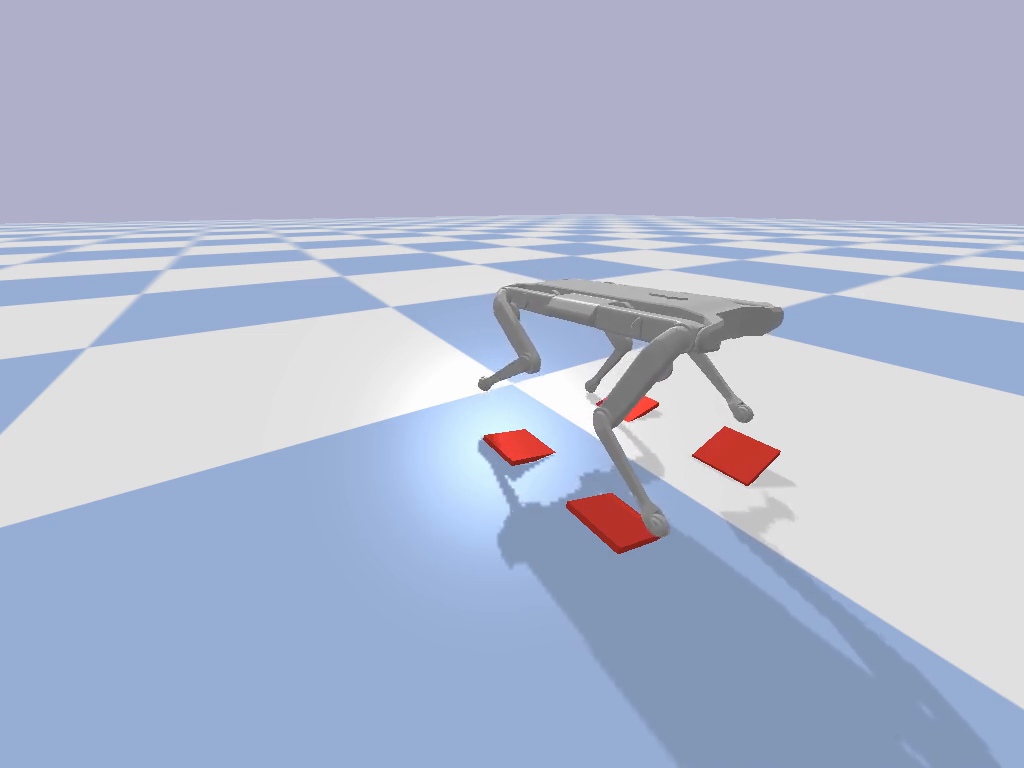}
    \end{minipage}
    \,\,\,
    \begin{minipage}[b]{.1\columnwidth}
    \centering
    \includegraphics[trim=400 100 150 250 ,clip,scale=0.09]{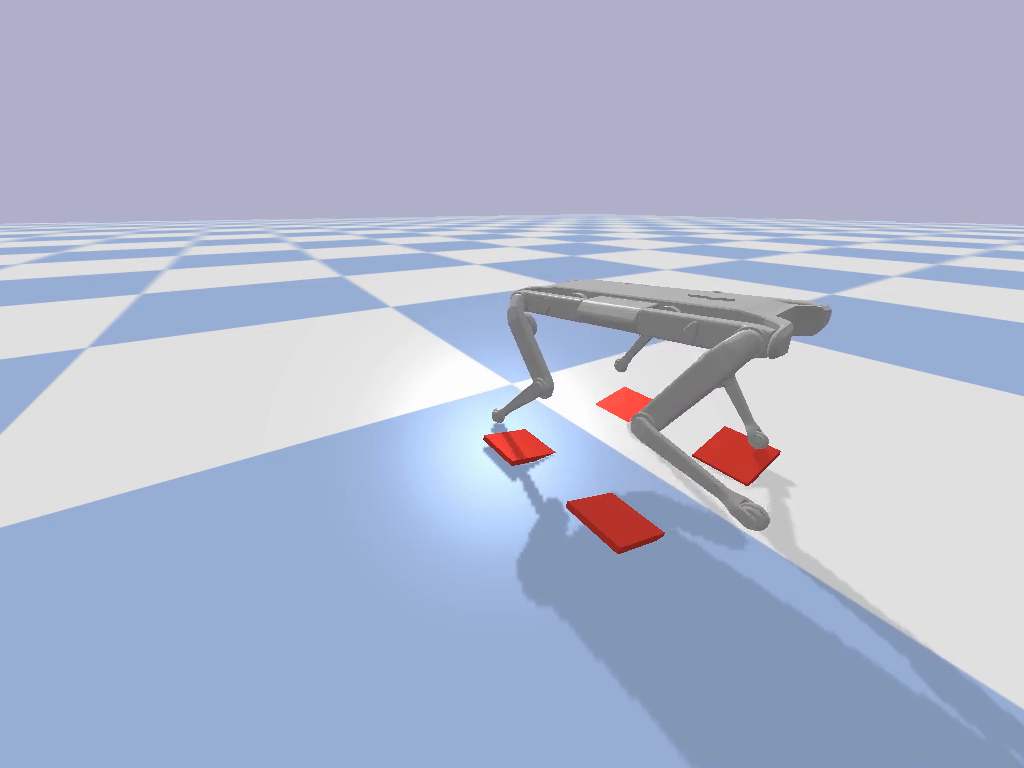}
    \end{minipage}
    \,\,\,
    \begin{minipage}[b]{.1\columnwidth}
    \centering
    \includegraphics[trim=400 100 150 250 ,clip,scale=0.09]{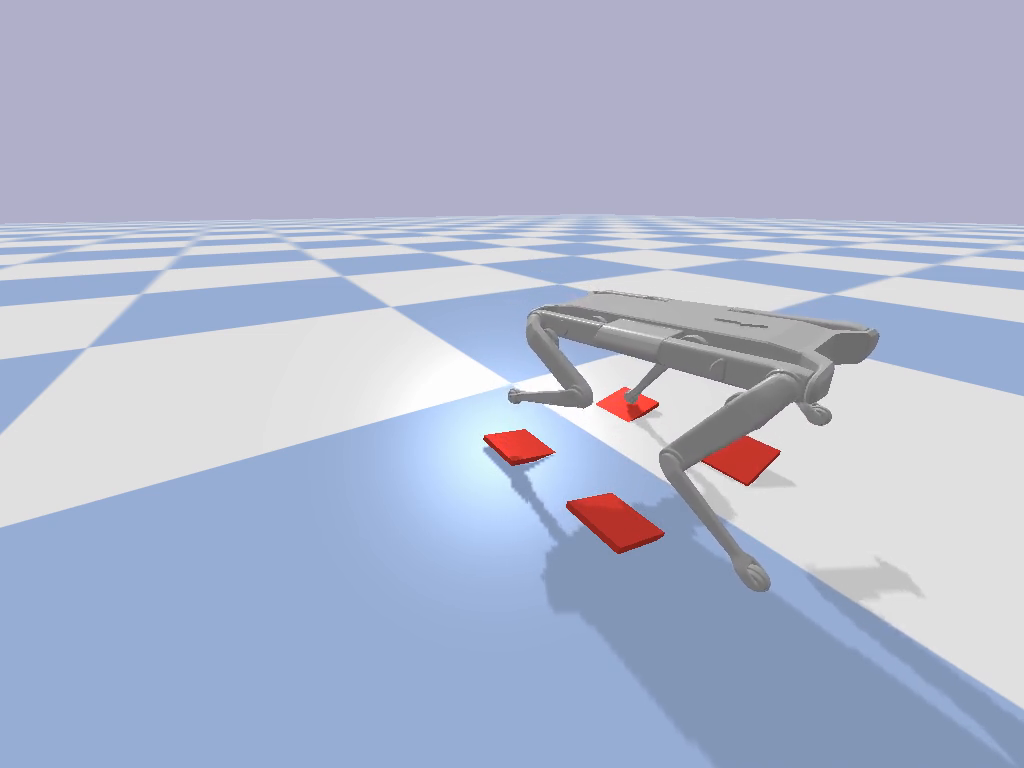}
    \end{minipage}
    \,\,\,
    \caption{Trot motion in an unplanned cluttered environment using stochastic centroidal SCP and whole-body DDP.}
    \label{fig: trot snapshots}
\end{figure}
\begin{table}[!t]
\centering
\caption{Whole-body DDP cost weights.}
\begin{tabular}{cccccc}
\toprule
\multicolumn{1}{c}{} & \multicolumn{2}{c}{\textbf{DDP solver task weights}}  \\
\cmidrule(rl){2-3} 
\textbf{Task} & {Trot} & {Bound}  \\
\midrule
Swing foot & 1e6 & 1e6\\
SCP CoM tracking & 1e3 & 1e1\\
SCP centroidal tracking & 1e3 & 1e3\\
SCP force tracking & 1e2 & 8e1\\
Friction cone & 2e2 & 2e0\\
State regulation & 1e-1 & 1e-1 \\
Control regulation & 1e0 & 1e0  \\
Contact impact velocity regulation & 2e1 & 2e1\\
\bottomrule
\end{tabular}
\label{table:DDP parameters}
\end{table}

During simulation (i.e. online), whole-body DDP joint-space trajectories were tracked at a higher sampling rate of $\Delta_k = 1$ ms using a PD control law: 
\begin{equation}
\label{eq:joint-space control}
\boldsymbol{\tau}_k = \Bar{\boldsymbol{\tau}}_k + \boldsymbol{K}_p (\boldsymbol{q}_k - \tilde{\boldsymbol{q}}_k) + \boldsymbol{K}_d (\dot{\boldsymbol{q}}_k - \dot{\tilde{\boldsymbol{q}}}_k),
\end{equation}
where $\bar{\boldsymbol{\tau}}_k$ are the DDP optimal feedforward joint torque controls, $\boldsymbol{q}_k$ and $\dot{\boldsymbol{q}}_k$ represent the DDP optimal joint positions and velocities respectively. Although in theory the optimal DDP gains could be used, it was not transferable in our case for highly dynamic motions, especially with long horizon as the motion diverged quickly. For that reason, we used hand-tuned PD gains for the scenarios described in the following subsection. 
The chance-constraints hyper parameters of the stochastic SCP were tuned as follows for the trotting and bounding motions: the probability level of friction pyramid constraint violations for every leg is $\alpha_u = 0.1$. The covariance of the contact position parametric uncertainties for each foot is set to  $\boldsymbol{\Sigma}_\theta  =\text{diag}\begin{bmatrix} 0.4^2,0.4^2,0.4^2 \end{bmatrix}$. The covariance of additive centroidal uncertainties is set to $\boldsymbol{\Sigma}_w = \text{diag}\, \begin{bmatrix} 0.85^2, 0.4^2, 0.01^2, 0.75^2, 0.4^2, 0.01^2, 0.85^2, 0.4^2, 0.01^2\end{bmatrix}$, and $\boldsymbol{\Sigma}_w =\text{diag}\, \begin{bmatrix}
0.75^2, 0.4^2,0.01^2,0.85^2,0.4^2,0.01^2, 0.75^2, 0.4^2, 0.01^2\end{bmatrix}$ for the trotting and bounding motions, respectively.
\subsection{Simulations setup}
\begin{figure}[!t]
    \begin{minipage}[b]{.1\columnwidth}
    \centering
    \includegraphics[trim=400 100 50 250 ,clip,scale=0.07]{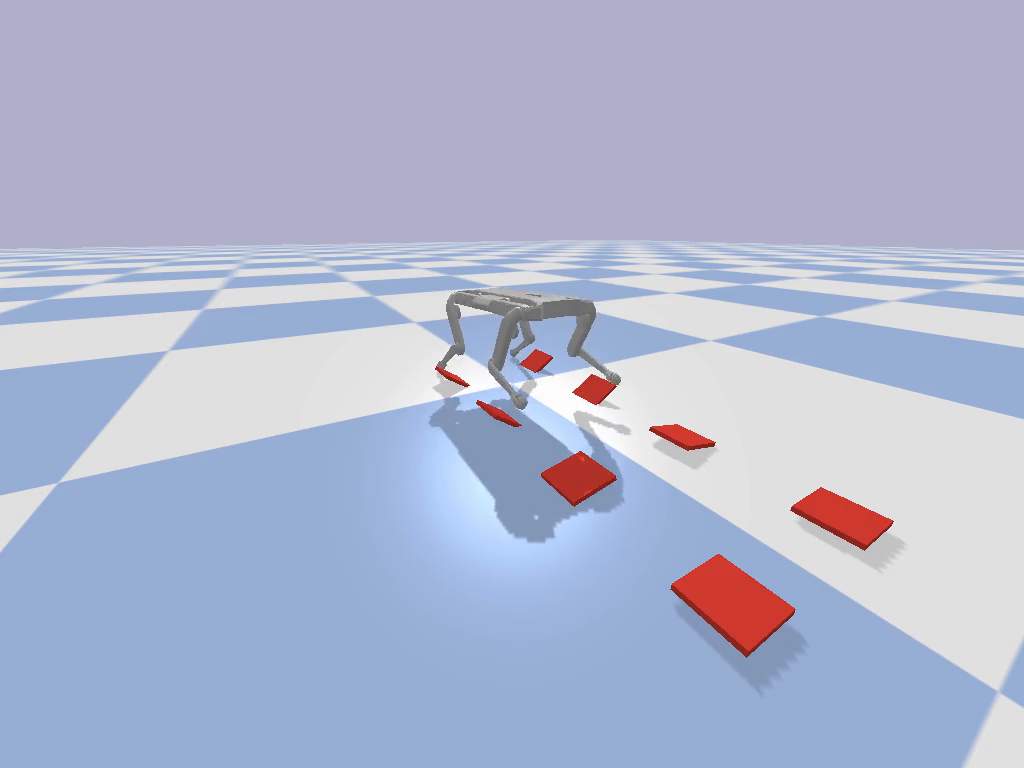}
    \end{minipage}
    \,
    \begin{minipage}[b]{.1\columnwidth}
    \includegraphics[trim=400 100 50 250 ,clip,scale=0.07]{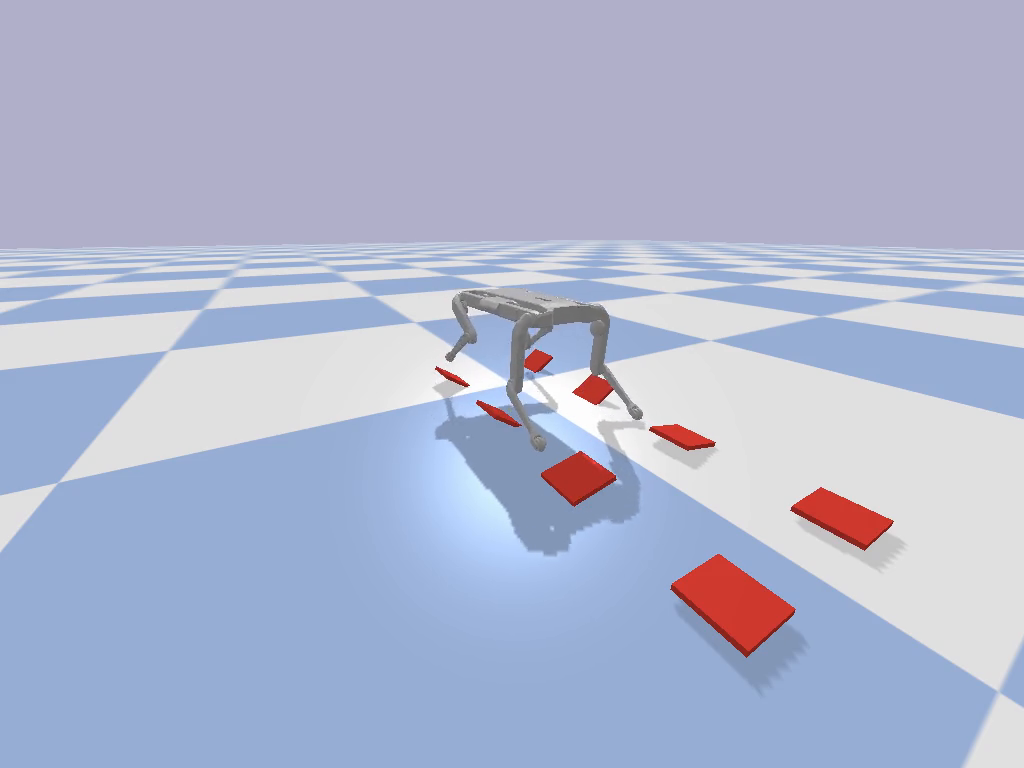}
    \end{minipage}
    \,
    \begin{minipage}[b]{.1\columnwidth}
    \centering
    \includegraphics[trim=400 100 50 250 ,clip,scale=0.07]{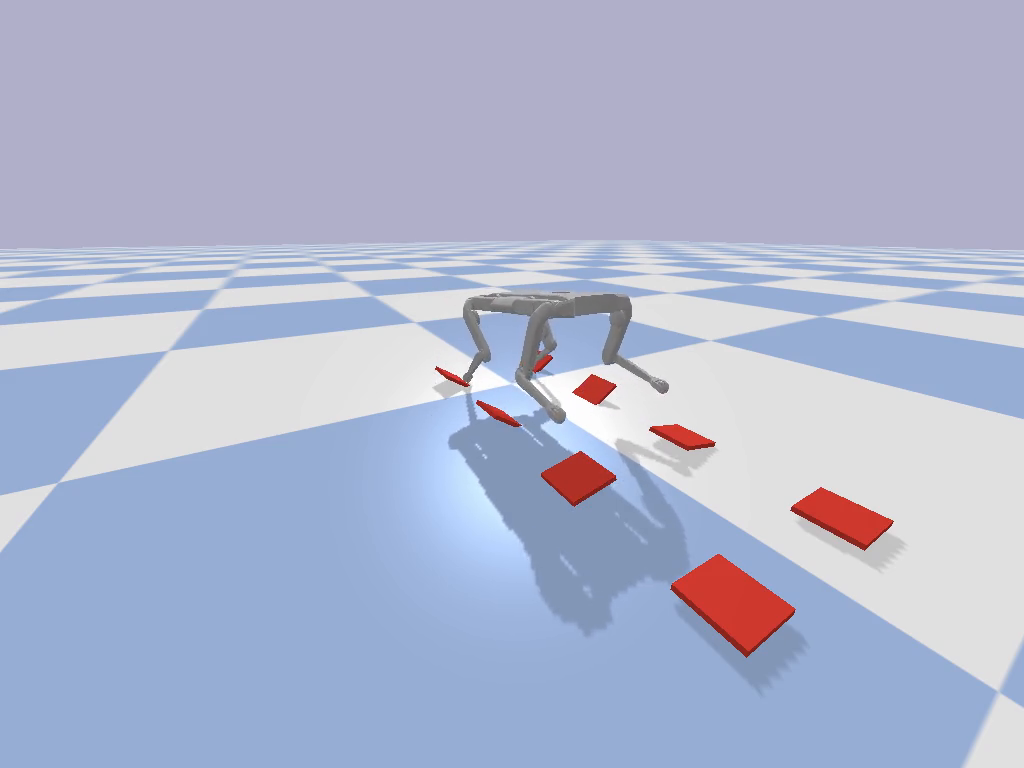}
    \end{minipage}
    \,
    \begin{minipage}[b]{.1\columnwidth}
    \centering
    \includegraphics[trim=400 100 50 250 ,clip,scale=0.07]{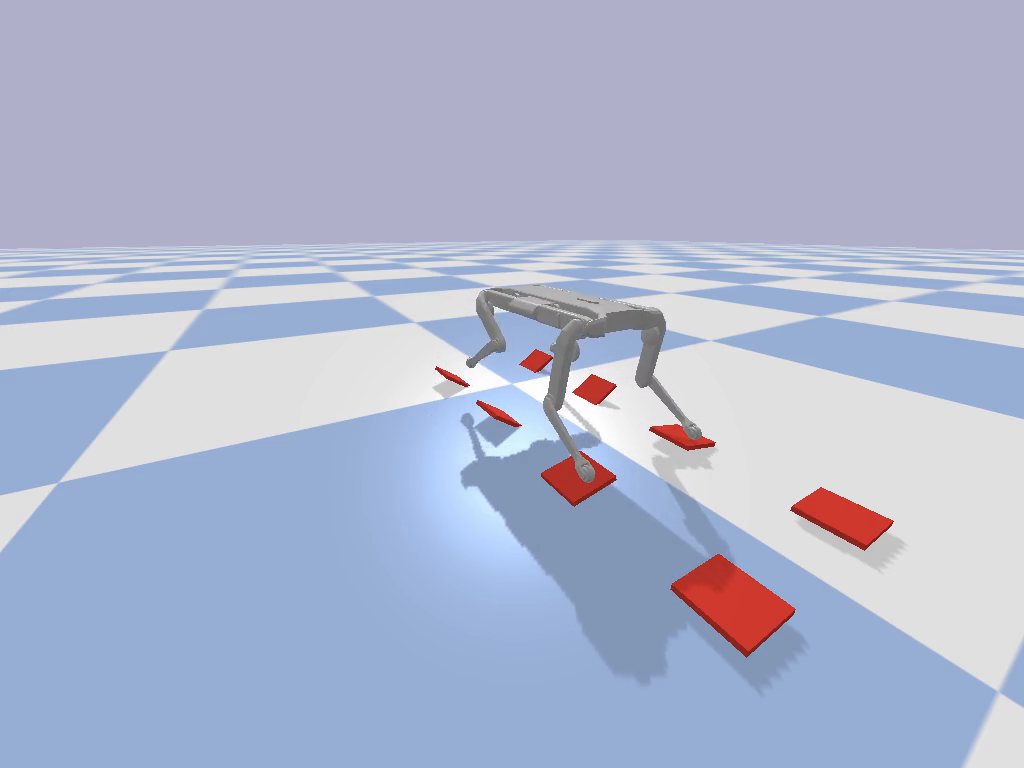}
    \end{minipage}
    \,
    \begin{minipage}[b]{.1\columnwidth}
    \centering
    \includegraphics[trim=400 100 50 250 ,clip,scale=0.07]{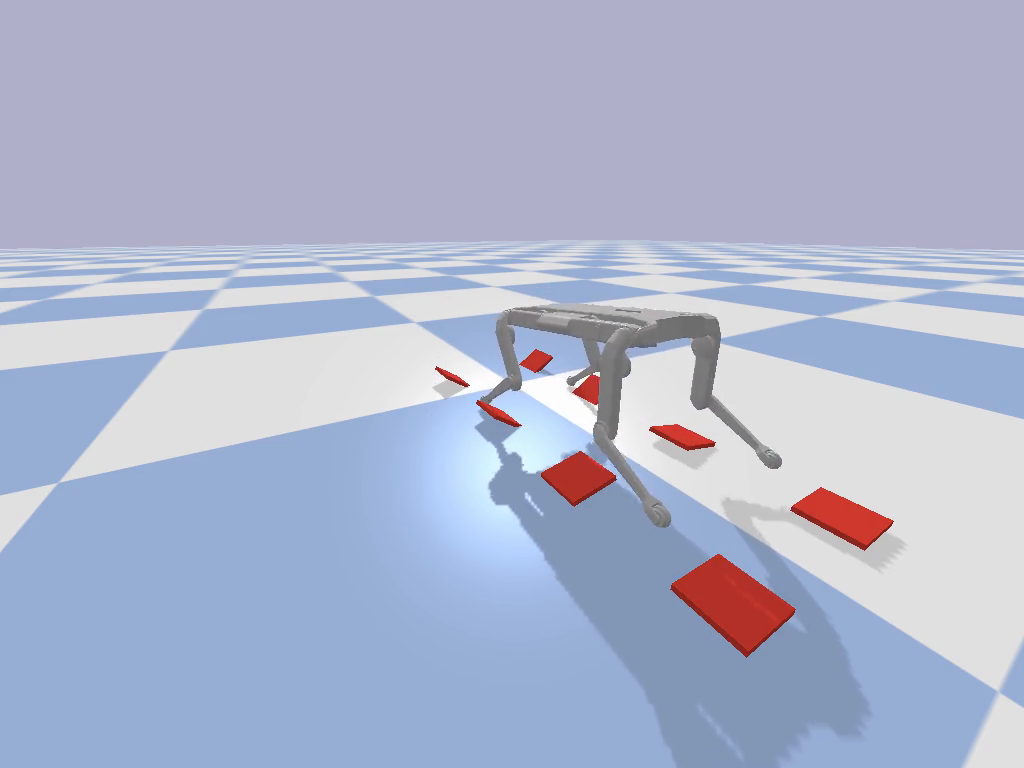}
    \end{minipage}
    \,
    \begin{minipage}[b]{.1\columnwidth}
    \includegraphics[trim=400 100 50 250 ,clip,scale=0.07]{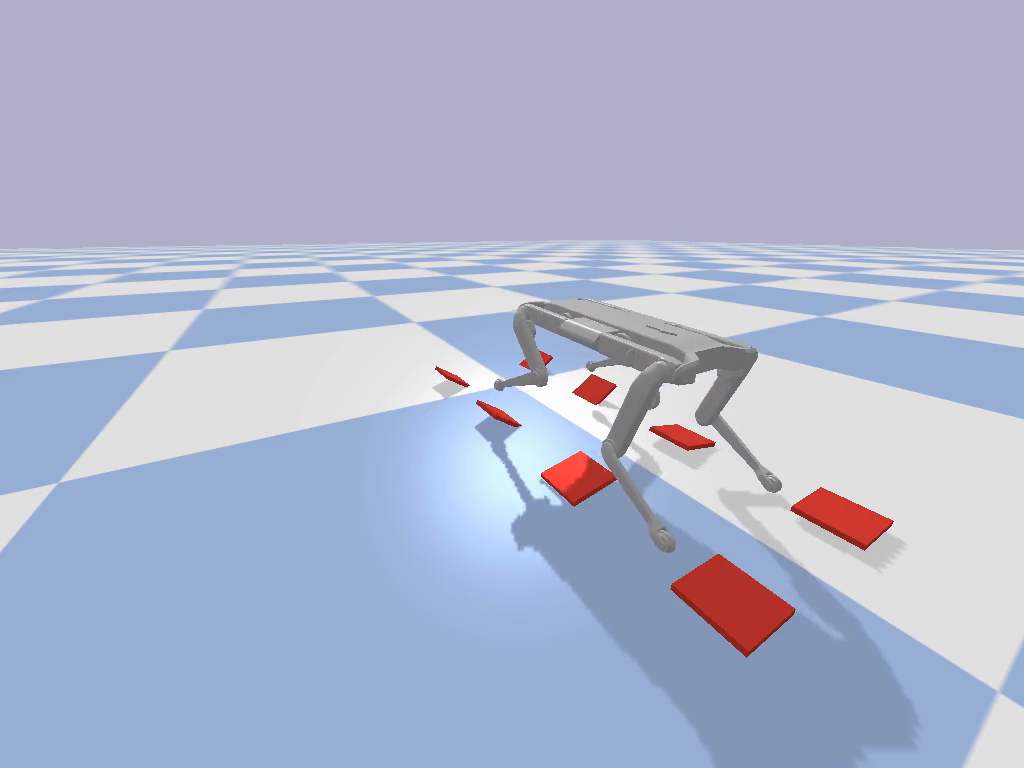}
    \end{minipage}
    \,
    \begin{minipage}[b]{.1\columnwidth}
    \centering
    \includegraphics[trim=400 100 50 250 ,clip,scale=0.07]{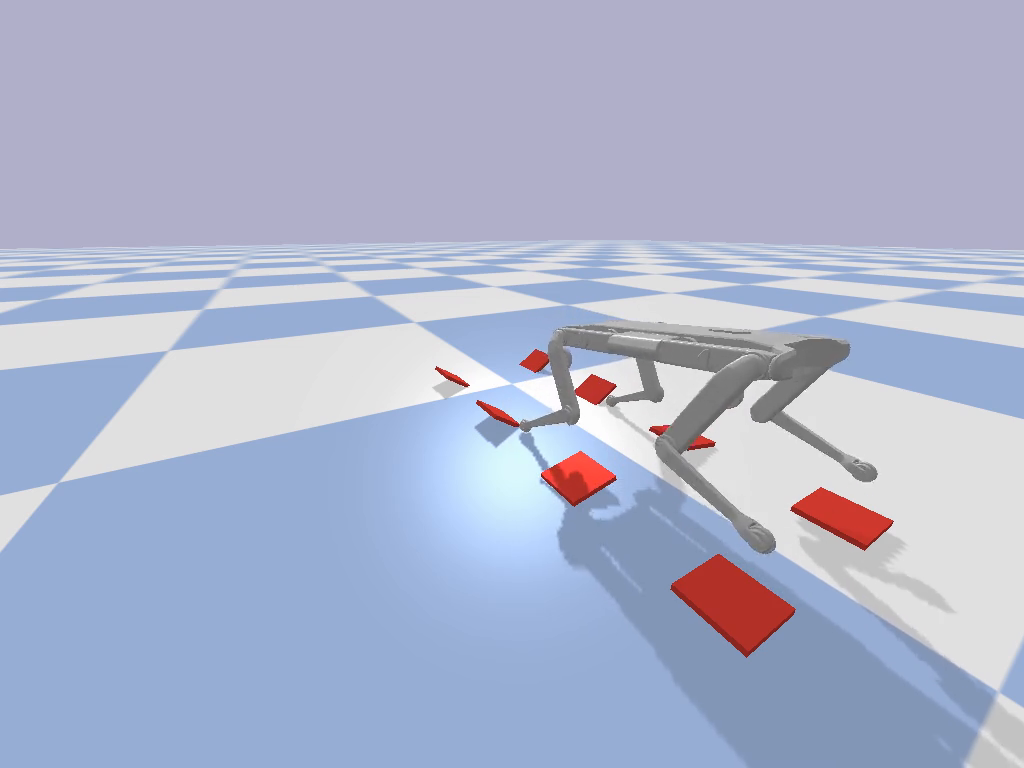}
    \end{minipage}
    \,
    \begin{minipage}[b]{.1\columnwidth}
    \centering
    \includegraphics[trim=400 100 50 250 ,clip,scale=0.07]{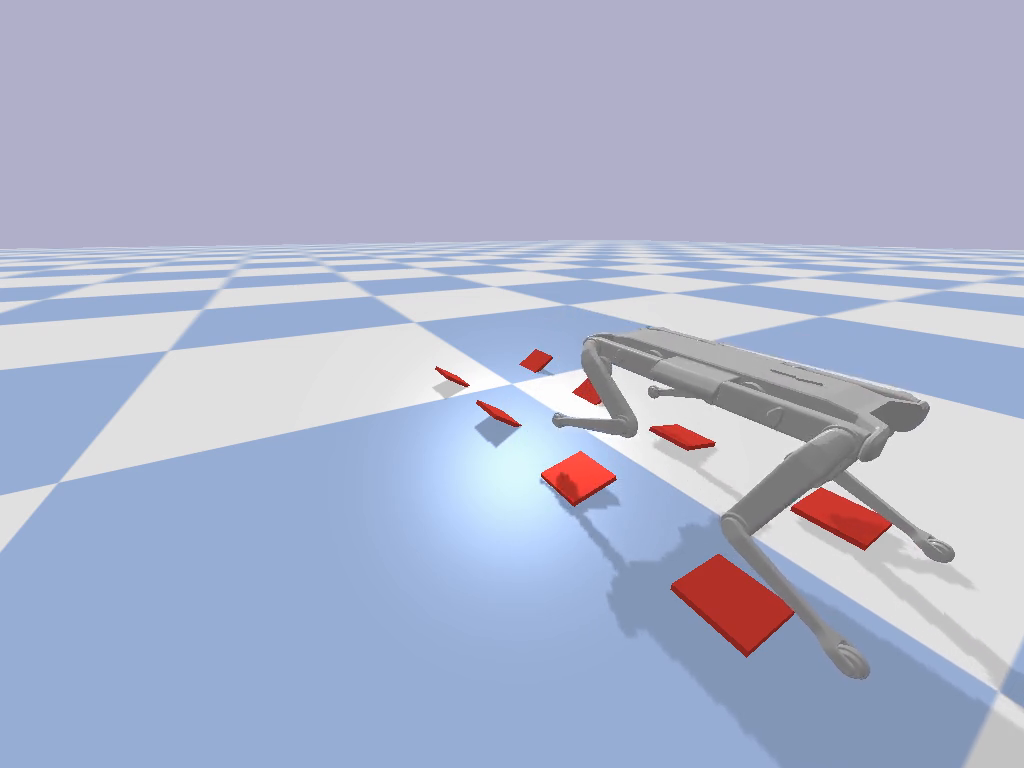}
    \end{minipage}
    \,
    \caption{Bound motion in an unplanned cluttered environment using stochastic centroidal SCP and whole-body DDP.}
    \label{fig: bound snapshots}
\end{figure}

 We ran a set of Monte-carlo simulations for two scenarios per motion: Scenario 1) \textbf{without debris}: 100 simulations on flat ground with a reduced floor friction $\mu=0.4$, while applying random lateral force disturbances for 200 ms at the center of the robot's base link. For trotting motion, we set $\boldsymbol{K}_p = 4.0*\mathbb{I}$, $\boldsymbol{K}_d = 0.2*\mathbb{I}$. For the bounding motion, we set $\boldsymbol{K}_p = 3.0*\mathbb{I}$, $\boldsymbol{K}_d = 0.2*\mathbb{I}$. Scenario 2) \textbf{with debris}: 100 simulations with reduced floor friction $\mu=0.4$, while adding unplanned debris of $2-3$ cm height ($6.6-10\%$ of the robot's leg length) with varying orientations of $0-17$ degrees along x and y directions as shown in Fig.~\ref{fig: trot snapshots} and Fig.~\ref{fig: bound snapshots} for trotting and bounding motions, respectively (please refer to the \href{https://www.youtube.com/watch?v=IZPyJZe7UDc}{video} for more details).The joint impedances were set to $\boldsymbol{K}_p = 5.0*\mathbb{I}$, $\boldsymbol{K}_d = 0.2*\mathbb{I}$ for the trotting motion, and $\boldsymbol{K}_p = 4.7*\mathbb{I}$, $\boldsymbol{K}_d = 0.2*\mathbb{I}$ for the bounding motion. Further, we apply again random lateral force impulses for 200 ms at the center of the robot's base. 
\begin{table}[!t]
\centering
\caption{Centroidal SCP cost weights.}
\begin{tabular}{cccccc}
\toprule
\multicolumn{1}{c}{} & \multicolumn{2}{c}{\textbf{SCP solver task weights}}  \\
\cmidrule(rl){2-3} 
\textbf{Task} & {Trot} & {Bound}  \\
\midrule
DDP CoM tracking & 1e4 & 1e4\\
DDP linear momentum tracking & 1e3 & 1e3\\
DDP angular momentum tracking & 1e5 & 1e5\\
Lateral force regulation per foot (x-direction) & 1e2 & 1e2 \\
Lateral force regulation per foot (y-direction) & 1e0 & 1e2 \\
Vertical force regulation per foot & 1e1 & 1e1  \\
Initial trust region weight & 1e2 & 1e2  \\
\bottomrule
\end{tabular}
\label{table:scp parameters}
\end{table}

 The same force disturbances were applied to the nominal and stochastic trajectories, and were sampled from a Gaussian distribution with zero mean and $\sigma = 15$ N ($60 \%$ of the robot weight). The force impulse is applied at the same randomly sampled time instance after the first second of the motion. We analyze the robustness of the motions generated using stochastic SCP against their nominal counterpart by evaluating the Normalized cumulative sum of the contact position deviations of the robot feet when a foot is in contact with the ground (i.e. foot slippage), which reflects the saturation of the friction pyramid constraints. The normalized cumulative sum was computed by subtracting the average cumulative sum of the previous samples from the current integral quantity at each point in time. Moreover, we report the centroidal tracking performance between the generated SCP references and the simulated trajectories. 

\begin{figure}[!t]
    \begin{minipage}[b]{.5\columnwidth}
    \centering
    \includegraphics[scale=0.38]{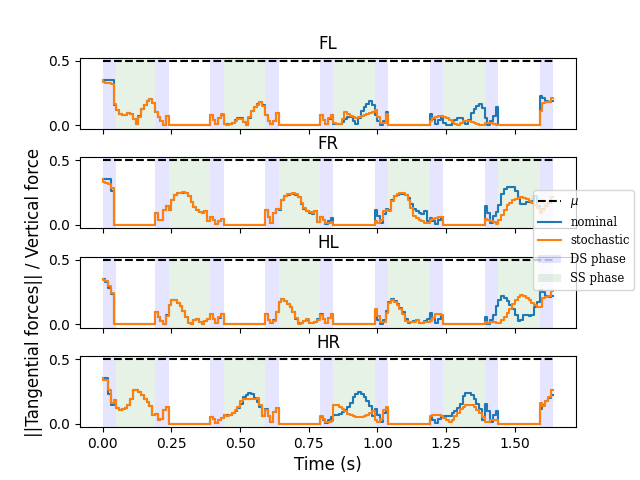}
    \end{minipage}
    \begin{minipage}[b]{.5\columnwidth}
    \centering
    \includegraphics[scale=0.38]{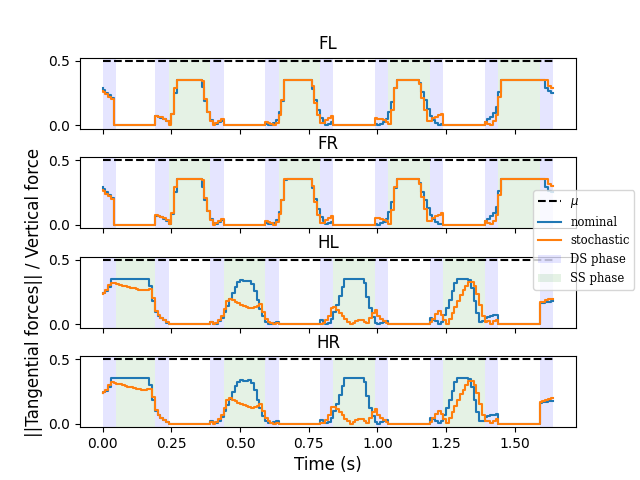}
    \end{minipage}
    \caption{Ratio of norm of tangential forces w.r.t. vertical force for a trotting motion (left) and a bounding motion (right).}
    \label{fig:SCP forces}
\end{figure}

First, we discuss the optimized contact forces generated using nominal and stochastic SCPs, which are later tracked using whole-body DDP. In Fig.~\ref{fig:SCP forces}, we plot the ratio between the norm of the tangential forces and the vertical forces for dynamic trotting and bounding motions. As expected, the forces optimized using stochastic SCP saturate less the friction cones compared to the ones optimized with nominal SCP, especially during single support phases where the solution of the QP is unique. This highlights the contribution of the control back-off magnitudes, which increase along the horizon due to the covariance propagation along the linearized dynamics~\eqref{eq:linearized control chance constraints}. 

\begin{figure}[!t]
    \begin{minipage}[b]{.5\columnwidth}
    \centering
    \includegraphics[scale=0.38]{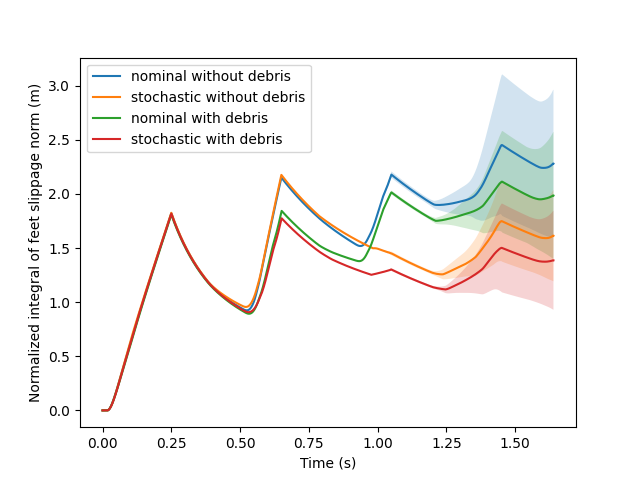}
    \end{minipage}
    \begin{minipage}[b]{.5\columnwidth}
    \centering
    \includegraphics[scale=0.38]{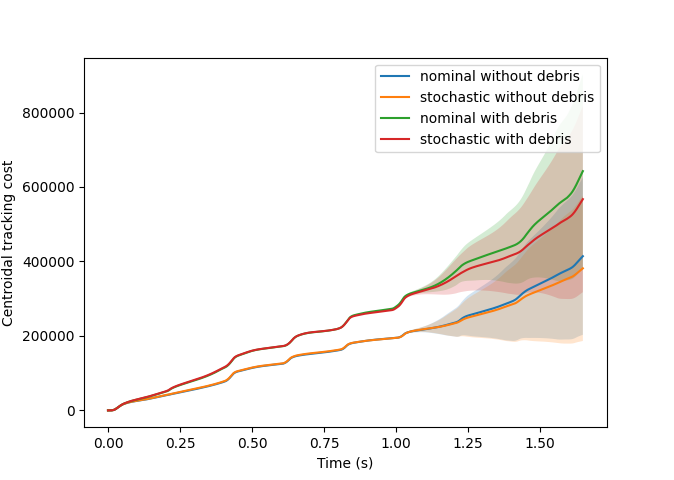}
    \end{minipage}
    \caption{Normalized cumulative sum of feet slippage norm (left) and centroidal tracking cost (right) for a trotting motion.}
    \label{fig:simulations trot}
\end{figure}

\begin{figure}[!t]
    \begin{minipage}[b]{.5\columnwidth}
    \centering
    \includegraphics[scale=0.38]{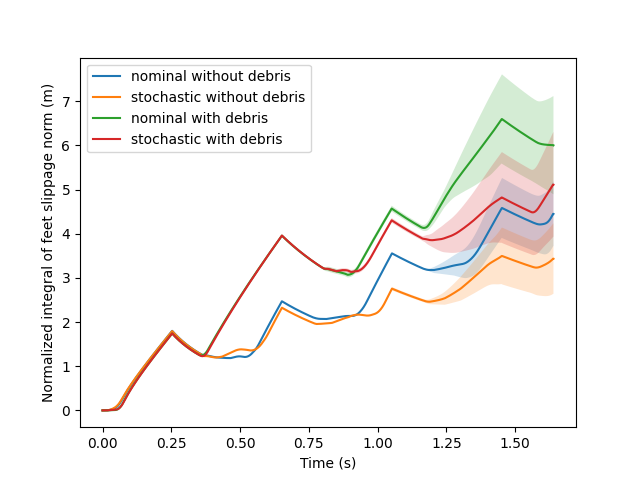}
    \end{minipage}
    \quad
    \begin{minipage}[b]{.5\columnwidth}
    \centering
    \includegraphics[scale=0.38]{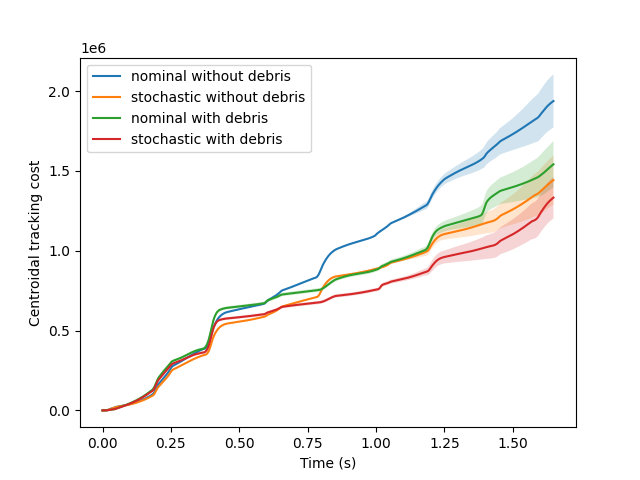}
    \end{minipage}
    \caption{Normalized cumulative sum of feet slippage norm (left) and centroidal tracking cost (right) for a bounding motion.}
    \label{fig:simulations bound}
\end{figure}

For the trotting motion (Fig.~\ref{fig:simulations trot}), trajectories designed using stochastic SCP (our method) achieved less feet slippage mean ($26.3 \%$ and $28.9\%$ for scenario 1 and 2, respectively) than nominal SCP and an improved centroidal tracking performance mean ($8.41 \%$ and $13.0\%$). The same analysis was carried out for the bounding motion in a more challenging terrain (Fig.~\ref{fig: bound snapshots}). As shown in Fig.~\ref{fig:simulations bound}, stochastic SCP trajectories contributed to less feet slippage mean ($22.8 \%$ and $14.8 \%$ for scenarios 1 and 2, respectively) than nominal SCP, and an improved centroidal tracking performance mean ($25.6 \%$ and $13.6 \%$).

\section{Discussion and Conclusions}
In this work, we used nonlinear stochastic trajectory optimization for generating robust centroidal momentum trajectories for legged robots that take into account additive uncertainties on the centroidal dynamics as well as parametric uncertainties on the contact positions. We used a linearization-based covariance propagation for resolving the stochastic nonlinear dynamics. Furthermore, we resolved the friction pyramid joint chance constraints by designing proper upper bounds (back-offs) at each point in time on the individual hyper-planes forming the friction pyramid polytopes. Finally, we presented a whole-body trajectory optimization framework that alternates between stochastic centroidal trajectory optimization and whole-body trajectory optimization for generating feasible robust whole-body motions. We used our framework to generate trotting and bounding dynamic gaits for the quadruped robot Solo. We then tracked these trajectories in a Pybullet physics simulator, introducing different disturbance realizations and contact uncertainties. The results show that our approach generated safer motions by contributing to less average contact slippage, as well as improved centroidal tracking performance over deterministic trajectory optimization. Although the current stochastic SCP approach does not require additional optimization variables over a deterministic approach SCP, the computational complexity is relatively higher due to uncertainty propagation and the additional tensor derivatives required for solving the linearized chance-constraints. 

Another limitation of the current stochastic SCP approach lies in the accuracy of uncertainties propagation through the linearized dynamics, which might be hindered for long horizons. However, we believe that this might not be an issue in practice when applied in receding horizon fashion. To this end, we plan to extend the current framework to nonlinear stochastic MPC in the future.  

\bibliographystyle{bibtex/splncs03_unsrt}
\bibliography{Biblio}
\end{document}